*Article*

# A Method for Estimating the Entropy of Time Series Using Artificial Neural Networks

Andrei Velichko [1,*] and Hanif Heidari [2,*]

[1] Institute of Physics and Technology, Petrozavodsk State University, 185910 Petrozavodsk, Russia

[2] Department of Applied Mathematics, School of Mathematics and Computer Sciences, Damghan University, Damghan, 36715-364, Iran

* Correspondence: velichko@petrsu.ru (A.V.); hanif.heidari@gmail.com (H.H.); Tel.: +7-911-400-5773 (A.V.)

**Abstract:** Measuring the predictability and complexity of time series using entropy is essential tool designing and controlling a nonlinear system. However, the existing methods have some drawbacks related to the strong dependence of entropy on the parameters of the methods. To overcome these difficulties, this study proposes a new method for estimating the entropy of a time series using the LogNNet neural network model. The LogNNet reservoir matrix is filled with time series elements according to our algorithm. The accuracy of the classification of images from the MNIST-10 database is considered as the entropy measure and denoted by NNetEn. The novelty of entropy calculation is that the time series is involved in mixing the input information in the reservoir. Greater complexity in the time series leads to a higher classification accuracy and higher NNetEn values. We introduce a new time series characteristic called time series learning inertia that determines the learning rate of the neural network. The robustness and efficiency of the method is verified on chaotic, periodic, random, binary, and constant time series. The comparison of NNetEn with other methods of entropy estimation demonstrates that our method is more robust and accurate and can be widely used in practice.

**Keywords:** entropy; time series; neural network; classification; MNIST-10 database; LogNNet

## 1. Introduction

Measuring the regularity of dynamical systems is one of the hot topics in science and engineering. For example, it is used to investigate the health state in medical science [1,2], for real-time anomaly detection in dynamical networks [3], and for earthquake prediction [4]. Different statistical and mathematical methods are introduced to measure the degree of complexity in time series data, including the Kolmogorov complexity measure [5], the $C_1/C_2$ complexity measure [5], and entropy [6].

Entropy is a thermodynamics concept that measures the molecular disorder in a closed system. This concept is used in nonlinear dynamical systems to quantify the degree of complexity. Entropy is an interesting tool for analyzing time series, as it does not consider any constraints on the probability distribution [7]. Shannon entropy (ShEn) and conditional entropy (ConEn) are the basic measures used for evaluating entropy. ShEn and ConEn measure the amount of information and the rate of information generation, respectively [1]. Based on these measures, other entropy measures have been introduced for evaluating the complexity of time series. For example, Letellier used recurrence plots to estimate ShEn [8]. Permutation entropy (PerEn) is a popular entropy measure that investigates the permutation pattern in time series [9]. Pincus introduced the approximate entropy (ApEn) measure, which is commonly used in the literature [10]. Sample entropy (SaEn) is another entropy measure that was introduced by Richman and Moorman [11]. The ApEn and SaEn measures are based on ConEp. All these methods are based on probability distribution and have shortcomings, such as sensitivity in short-length time series [12], equality in time series [6], and a lack of information related to the sample differences in amplitude [9]. To overcome these difficulties, many researchers have attempted to modify their methods. For example, Azami and Escudero introduced fluctuation-based dispersion entropy to deal with the fluctuations of time series [1]. Letellier used recurrent plots to evaluate Shannon entropy in time series with noise contamination. Watt and Politi investigated the efficiency of the PE method and introduced modifications to speed up the convergence of the method [13]. Molavipour et al. used neural networks to approximate the probabilities in mutual information equations, which are based on ShEn [14]. Deng introduced Deng entropy [15,16], which is a generalization of Shannon entropy. Martinez-Garcia et al. applied deep recurrent neural networks to approximate the probability distribution of the system outputs [17].

We propose a new method for evaluating the complexity of a time series which has a completely different structure compared to the other methods. It computes entropy directly, without considering or approximating probability distributions. The proposed method is based on LogNNet, an artificial neural network model [18,19]. Velichko [18] showed a weak correlation between the classification accuracy of LogNNet and the Lyapunov exponent of the time series filling the reservoir. Subsequently, we found that the classification efficiency is proportional to the entropy of the time series [20], and this finding led to the development of the proposed method. LogNNet can be used for estimating the entropy of time series, as the transformation of inputs is carried out by the time series, and this affects the classification accuracy. A more complex transformation of the input information, performed by the time series in the reservoir part, results in a higher classification accuracy in LogNNet.

To determine entropy, the following main steps should be performed: the LogNNet reservoir matrix should be filled with elements of the studied time series, and then the network should be trained and tested using handwritten MNIST-10 digits in order to determine the classification accuracy. Accuracy is considered to be the entropy measure and denoted as NNetEn.

To validate the method, we used the well-known chaotic maps, including the logistic, sine, Planck, and Henon maps, as well as random, binary, and constant time series.

This model has advantages compared with the existing entropy-based methods, including the availability of one control parameter (the number of network learning epochs), the fact that scaling the time series by amplitude does not affect the value of entropy, and the fact that it can be used for a series of any length. The method has a simple software implementation, and it is available for download to users in the form of an "NNetEn calculator 1.0.0.0" application.

The scientific novelty of the presented method is a new approach to estimating the entropy of time series using neural networks.

The rest of the paper is organized as follows. Section 2 describes the structure of LogNNet and the methodology used for calculating entropy. In addition, a new time series characteristic, called time series learning inertia, is described. The numerical examples and results are presented in Section 3. Section 4 summarizes the study with a discussion and outlines future research ideas.

## 2. Methods

This study applies the LogNNet 784:25:10 model [18] to calculate the entropy value. The LogNNet 784:25:10 model was designed to recognize images of handwritten digits in the range of 0–9 taken from the MNIST-10 dataset. The database contains a training set of 60,000 images and a test set of 10,000 images. Each image has a size of 28 × 28 = 784 pixels and is presented in grayscale color.

### 2.1. LogNNet 784:25:10 Model for Entropy Calculation

The model consists of two parts (see Figure 1). The first part is the model reservoir, which uses the matrix $W_1$ to transform the input vector $Y$ into another vector $S_h$ ($P$ = 25). The second part of the model contains a single layer feedforward neural network that transforms vector $S_h$ into digits 0–9 at the output layer $S_{out}$.

The algorithm has the following steps.

Step 1. Loading time series $x_n$.

Step 2. Loading the MNIST-10 database and the T-pattern-3 pattern [18] for transforming input images into the vectors array $Y$.

Step 3. Initializing the initial values of weights and neurons. The initial weights $W_2$ are set to 0.5. The constant initial weights ensure repeatable entropy results.

Step 4. The reservoir matrix $W_1$ is constructed using the given time series (Section 2.2.).

Step 5. Calculating the coefficients for normalization.

Step 6. The number of training epochs is set.

Step 7. The training process of the LogNNet 784:25:10 network is performed on a training set. The weights of the matrix $W_2$ are trained. The learning rate of the back propagation method is set to 0.2.

Step 8. The testing process of the LogNNet 784:25:10 network is performed on a test set and classification accuracy is calculated.

Step 9. The value of NNetEn entropy is defined as follows:

$$\text{NNetEn} = \frac{\text{Classification accuracy}}{100\%}$$

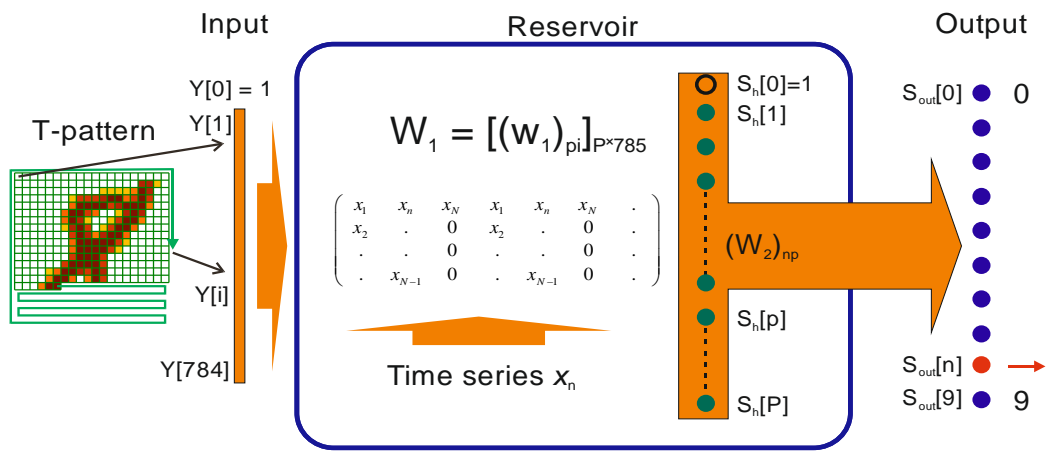

**Figure 1.** The LogNNet model structure.

Therefore, classification accuracy is considered to be the new entropy measure and is denoted as NNetEn (see Equation (1)).

Classification accuracy is distributed in the range from 0 to 100%, and NNetEn has values between 0 and 1.

## 2.2. Matrix Filling Methods

Matrix $W_1$ contains 25 rows and 785 columns and therefore requires 19,625 elements. However, in real-world problems, the length of the given time series can be either more or less than 19,625 elements.

Let us suppose that we have a time series with $N$ elements $x_n = (x_1, x_2, x_3, …, x_N)$.

If $N ≥ 19,625$, it is necessary to ignore $N$-19,625 elements of time series and fill the matrix with the remained 19,625 elements, filling the matrix column by column. We suggest ignoring the first elements, as this may be the transient period of a dynamic process. Many mathematical and physical systems have a transient period before the dynamics stabilize. This also holds for discrete chaotic maps (Section 3).

If $N < 19,625$, we propose a special method. This method implies the column-by-column filling of the matrix $W_1$. We set zero values for the remaining elements in a matrix row if the values are missing and a row cannot be filled with the elements of the given time series. A schematic description of the method is given in Figure 2a. As an example, a reduced matrix $W_1$ is filled with the series $x_n$ = (1, 2, 3, 4, 5, 6, 7, 8, 9) (Figure 2b).

$$\begin{pmatrix} x_1 & x_n & x_N & x_1 & x_n & x_N & . \\ x_2 & . & 0 & x_2 & . & 0 & . \\ . & . & 0 & . & . & 0 & . \\ . & x_{N-1} & 0 & . & x_{N-1} & 0 & . \end{pmatrix} \qquad \begin{pmatrix} 1 & 5 & 9 & 1 & 5 & 9 & 1 \\ 2 & 6 & 0 & 2 & 6 & 0 & 2 \\ 3 & 7 & 0 & 3 & 7 & 0 & 3 \\ 4 & 8 & 0 & 4 & 8 & 0 & 4 \end{pmatrix}$$

(a) \qquad\qquad (b)

**Figure 2.** The structure of the matrix filling method (**a**); an example of time series with 9 elements (**b**).

*2.3. Program Code for Calculating NNetEn*

For practical method application, the "NNetEn calculator 1.0.0.0" software was designed for the Windows operating system. It calculates the NNetEn time series recorded in a text file. The software interface is demonstrated in Figure 3. The code is implemented in the Delphi 7 programming language, and contains nine steps described in Section 2.1. The calculation of one time series with 100 epochs takes about 620s using one Intel ™ Co™TM) m3-7Y30 CPU @ 1.00GHz processor thread. Further description can be found in the Supplementary Materials of this paper.

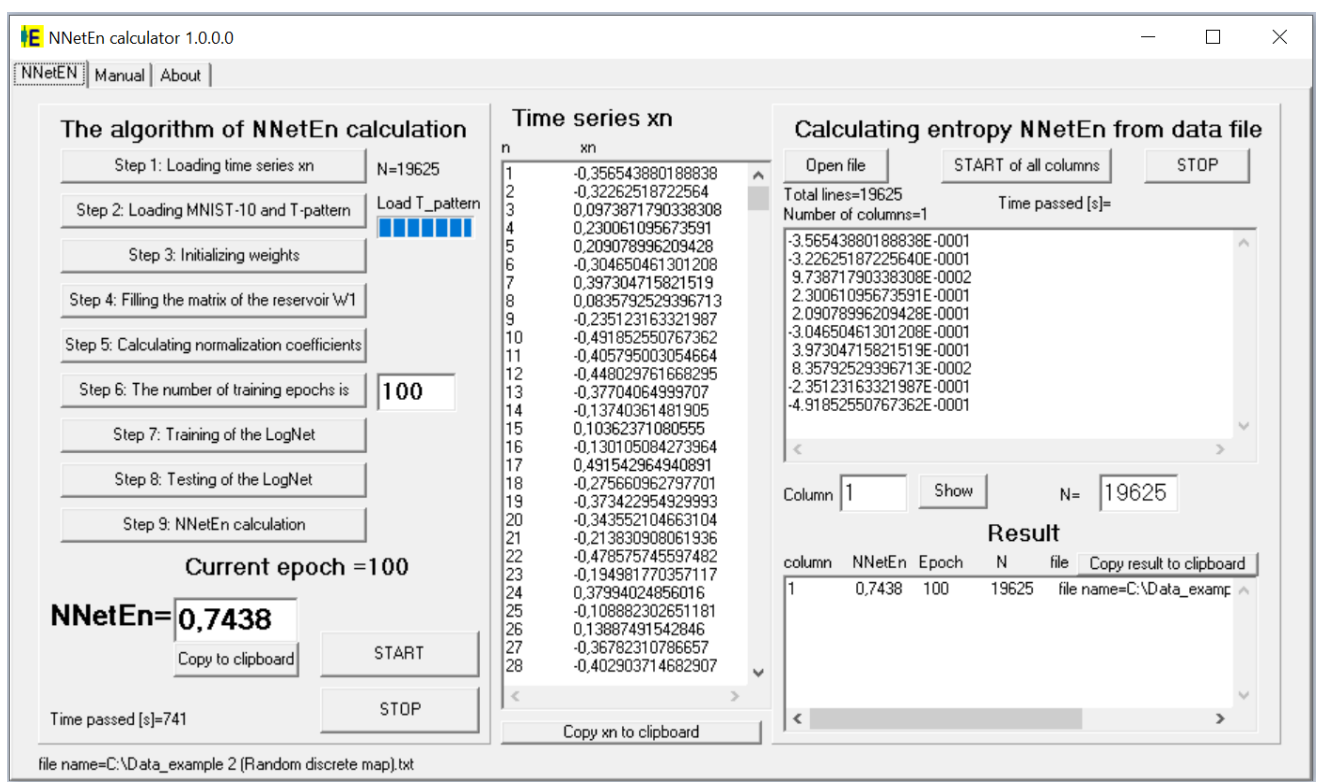

**Figure 3.** The interface of the "NNetEn calculator 1.0.0.0" program.

## 3. Results

To show the efficiency and robustness of the proposed method, we apply the method to chaotic, random, periodic, binary, and constant time series $x_n$ with different lengths.

I    Discrete chaotic maps:

Because of the transient period, the first 1000 elements are ignored. The NNetEn measure is calculated for $x_n$, where $n > 0$.

a) Logistic map:

$$x_{n+1} = r \cdot x_n \cdot (1 - x_n) \quad , \ 1 \leq r \leq 4, \ x_{-999} = 0.1, \tag{2}$$

b) Sine map:

$$x_{n+1} = r \cdot \sin(\pi \cdot x_n) \quad , \ 0.7 \leq r \leq 2, \ x_{-999} = 0.1, \tag{3}$$

c) Planck map:

$$x_{n+1} = \frac{r \cdot x_n^3}{1 + \exp(x_n)} \quad , \ 3 \leq r \leq 7, \ x_{-999} = 4 \tag{4}$$

d) Henon map:

$$\begin{cases} x_{n+1} = 1 - r \cdot x_n^2 + y_n \\ y_{n+1} = r_1 \cdot x_n \end{cases} , \ 0 \leq r \leq 1.4, \ r_1 = 0.3, \ x_{-999} = 0.1, \ y_{-999} = 0.1 \tag{5}$$

II    Random discrete map:

$$x_n = random - 0.5 \tag{6}$$

III    Periodic discrete map:

$$x_n = A \cdot \sin\left(\frac{n \cdot 20\pi}{19625}\right) \tag{7}$$

IV    Binary discrete map:

$$x_n = n \bmod 2 \tag{8}$$
$$x_n = (1, 0, 1, 0, 1, 0, 1, 0, 1, .....)$$

V    Constant discrete map:

$$x_n = A \tag{9}$$

### 3.1. Calculation of NNetEn Entropy Measure

In this subsection, we consider time series with $N = 19{,}625$ elements. To train the matrix $W_2$, we used the number of epochs to be equal to 100.

The dependence of NNetEn on the control parameter $r$ in the chaotic time series (Equations (2)–(5)) is presented in Figures 4–7.

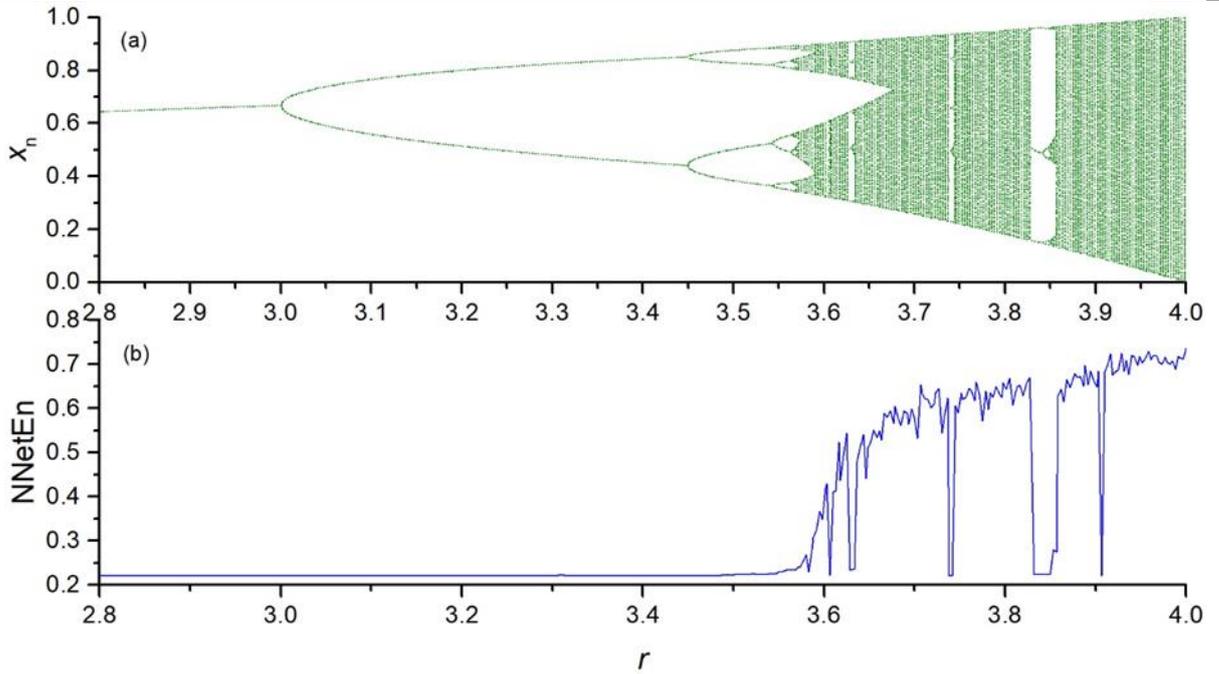

**Figure 4.** Bifurcation diagrams for logistic map (Equation (2)) (**a**); the dependence of NNetEn on the parameter *r* (number of epochs is 100) (**b**).

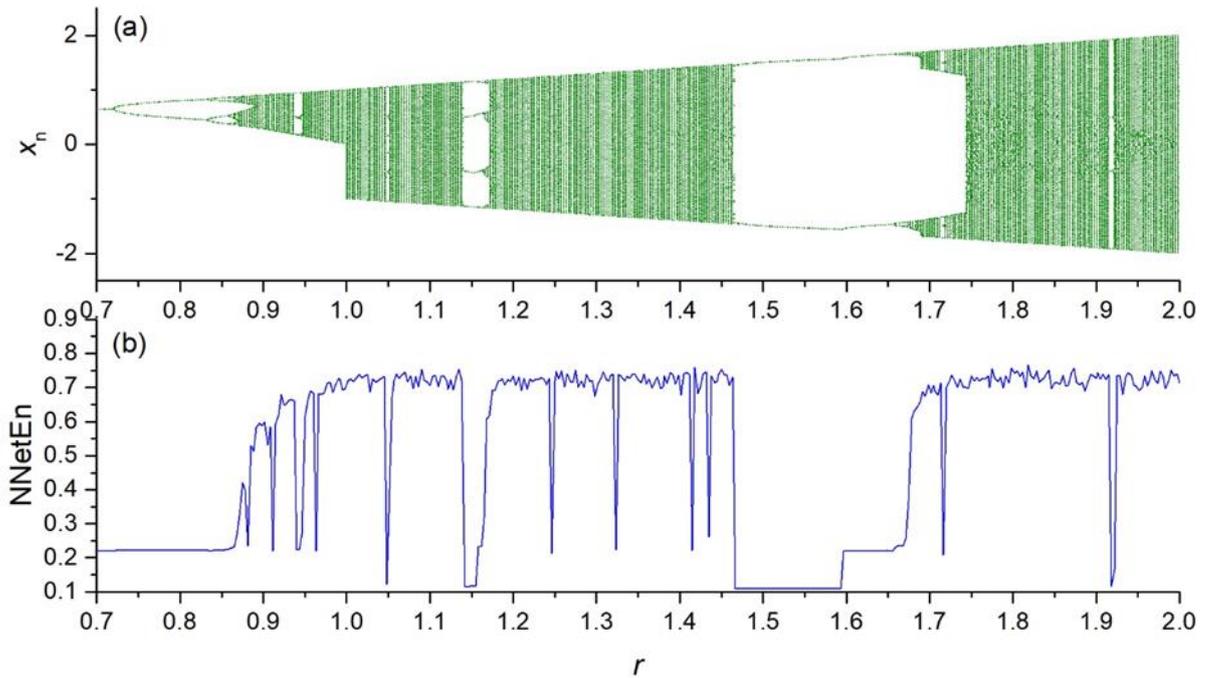

**Figure 5.** Bifurcation diagrams for sine map (Equation (3)) (**a**); the dependence of NNetEn on the parameter *r* (number of epochs is 100) (**b**).

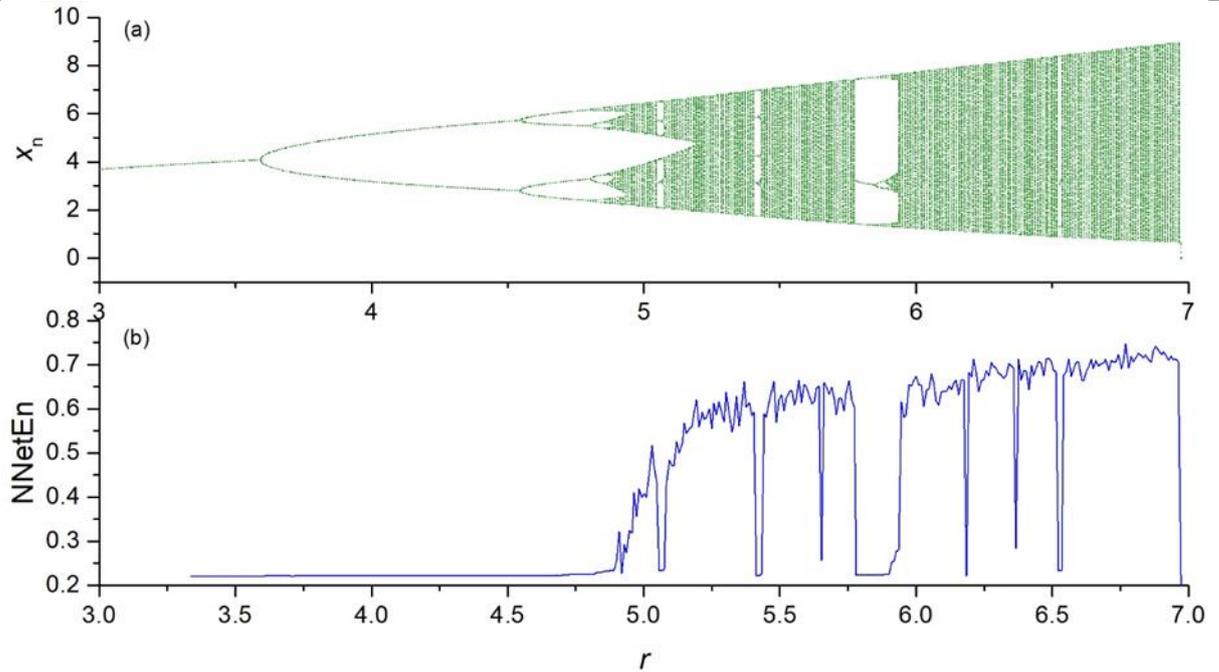

**Figure 6.** Bifurcation diagrams for Planck map (Equation (4)) (**a**); the dependence of NNetEn on the parameter $r$ (number of epochs is 100) (**b**).

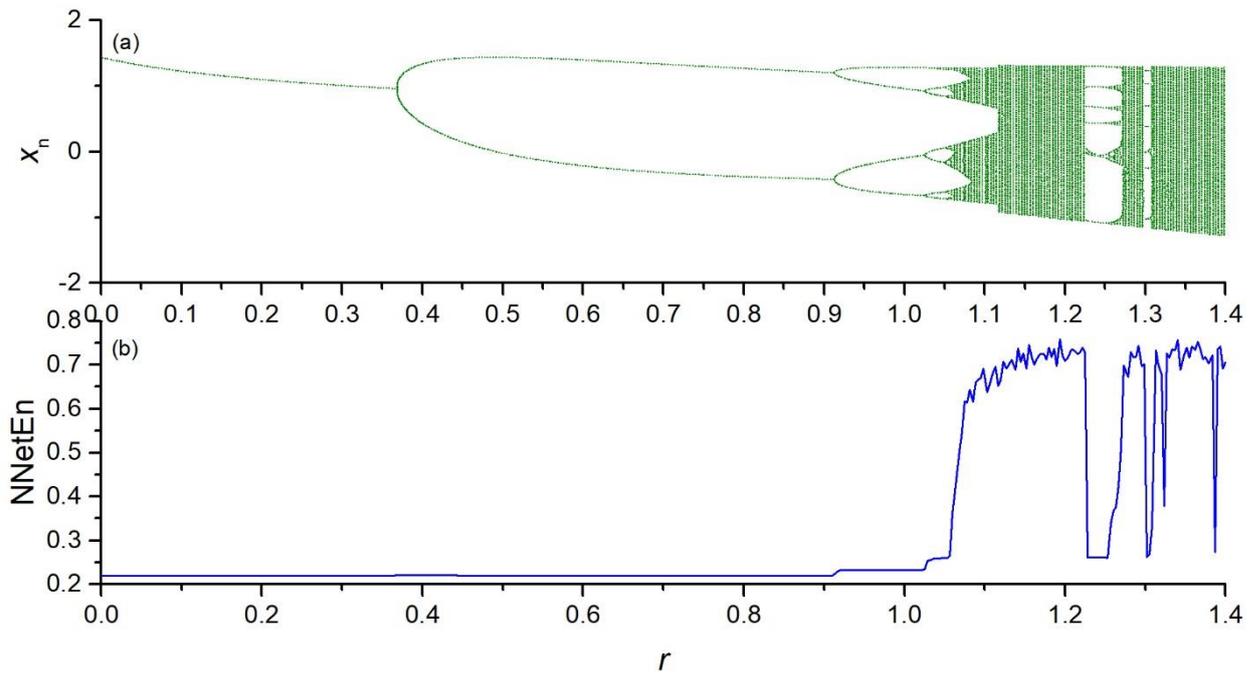

**Figure 7.** Bifurcation diagrams for Hennon map (Equation (5)) (**a**); the dependence of NNetEn on the parameter $r$ (number of epochs is 100) (**b**).

For the random time series (Equation (6)), the NNetEn equals 0.7438.
For the periodic time series (Equation (7)), the NNetEn for different values of the parameter $A$ is presented in Figure 8a.

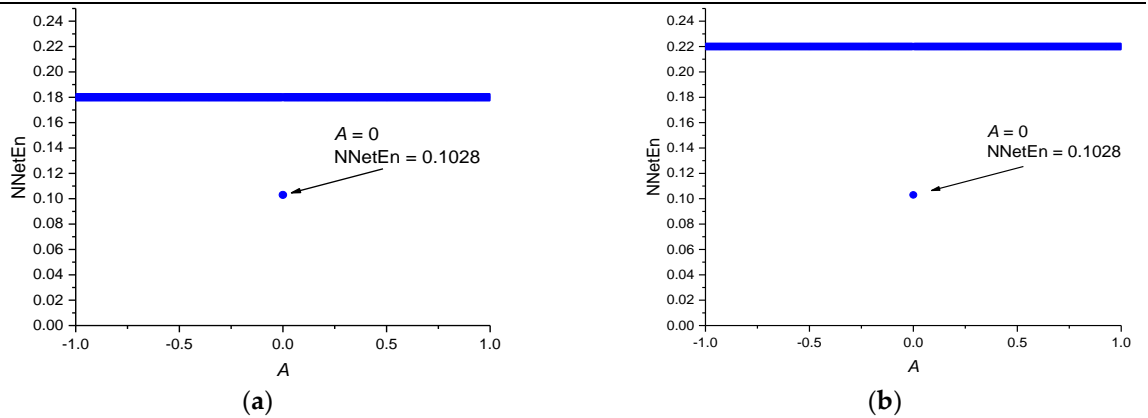

**Figure 8.** The dependence of NNetEn on the parameter *A* for the periodic time series (Equation (7)) (**a**); the dependence of NNetEn on the parameter *A* for constant time series (Equation (9)) (**b**).

For the binary series described by Equation (8), the NNetEn equals 0.2196.

The NNetEn values for constant time series are depicted in Figure 8b. Entropy has the same value NNetEn = 0.22 for $|A| > 0$ and NNetEn = 0.1028 for $A = 0$. Therefore, the lowest possible NNetEn value is about 0.1.

A comparison of the NNetEn values for chaotic, random, periodic, and constant time series demonstrates that the NNetEn increases when the complexity of the time series increases. Therefore, there is a direct relation between the degree of complexity and the NNetEn of time series. This confirms that NNetEn can be used for comparing the degree of complexity of a given time series. Another advantage of this method is that NNetEn is independent of signal amplitude *A*. The entropy of the signal should not depend on the multiplication of the entire time series by a constant.

*3.2. The Influence of the Number of Training Epochs on the NNetEn Value*

The influence of the number of epochs on the value of NNetEn was studied using a time series with $N = 19{,}625$ elements, generated by logistic mapping (Equation (2)). The results are presented in Figure 9a.

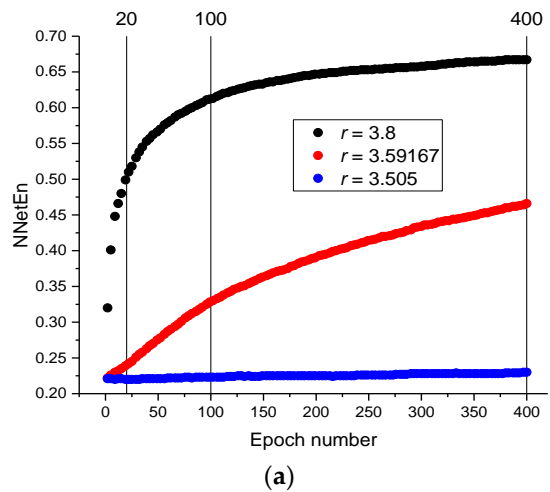

(**a**)

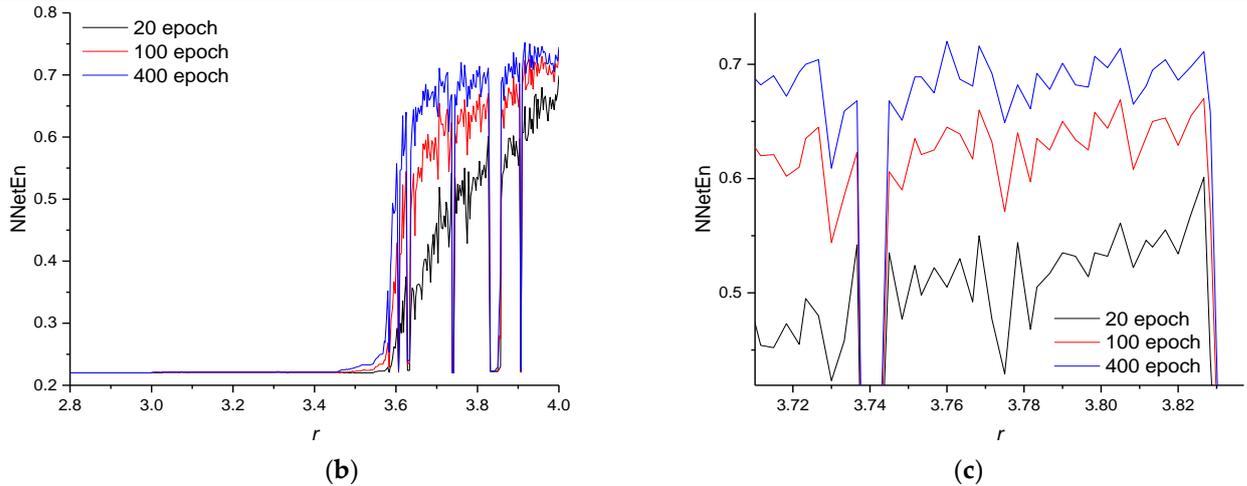

**Figure 9.** The relation between NNetEn and the number of epochs for the time series on logistic map (Equation (2)) (**a**); the dependence of NNetEn value on the parameter *r* using 20, 100, and 400 epochs (**b**); a magnification of subfigure (**b**) for *r* between 3.72 and 3.82 (**c**).

NNetEn gradually increases with an increasing number of epochs until a plateau is reached (Figure 9a). The speed of reaching the plateau depends on the type of signal. For example, the velocity of reaching the plateau at *r* = 3.59167 is slower than that at *r* = 3.8 and *r* = 3.505. Figure 9b shows the dependence of NNetEn on the parameter *r* for different numbers of epochs. The trends are similar, although there are differences in the details. The behaviors of NNetEn with 100 epochs and 400 epochs are more similar than NNetEn with 20 epochs and 400 epochs (Figure 9c). This example demonstrates that a significant number of epochs are required for reaching the plateau in NNetEn, especially in chaotic time series. Therefore, it is necessary to indicate the number of epochs as a parameter of the model.

### 3.3. Learning Inertia as a New Characteristic of Time Series

To identify the speed of the NNetEn convergence to the plateau with respect to the number of epochs, the parameter $\delta_{Ep1/Ep2}$ is proposed:

$$\delta_{Ep1/Ep2} = \frac{\text{NNetEn}(Ep2 \text{ epoch}) - \text{NNetEn}(Ep1 \text{ epoch})}{\text{NNetEn}(Ep2 \text{ epoch})}, \tag{10}$$

where $Ep1$ and $Ep2$ ($Ep1 < Ep2$) are the numbers of epochs used in calculating the entropy.

The parameter reflects the rate of change in NNetEn values when the number of epochs is decreased from $Ep2$ to $Ep1$. Figure 10 demonstrates the dependence of $\delta_{100/400}$ on the parameter *r* in the logistic map (Equation (2)). The maximum $\delta_{100/400}$ occurs at *r* = 3.59167. This means that the neural network has the lowest learning rate at *r* = 3.59167.

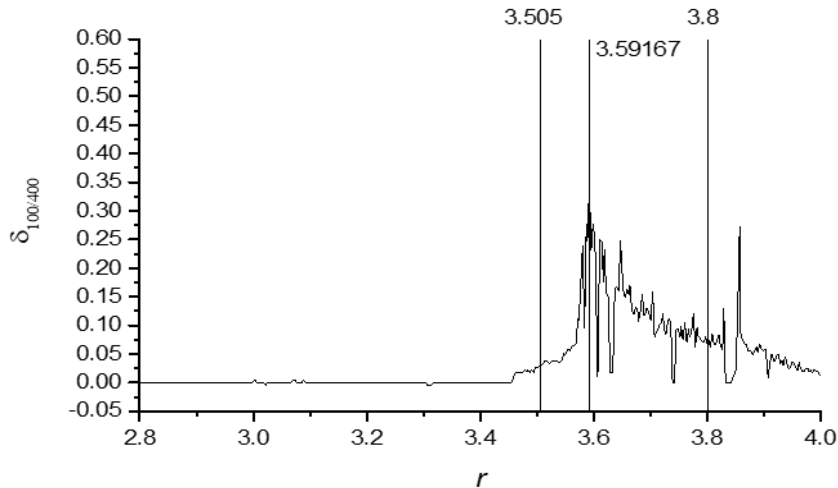

**Figure 10.** The parameter $\delta_{100/400}$ (learning inertia) in relation to the parameter *r*.

The parameter $\delta_{Ep1/Ep2}$ can be considered a new characteristic of the input time series and is named "learning inertia".

*3.4. Calculation of NNetEn Entropy with Variation in the Length of the Time Series N.*

Data obtained from financial markets [21], physical experiments [12,22], and biological or medical data [11,23] may have $N < 19{,}625$ elements. To investigate the efficiency and stability of the matrix filling method (see Section 2.2.), the method was applied to different types of time series with different *N*. The epoch number is set to 100 in this subsection.

The NNetEn values for time series based on the logistic map with $r = 3.8$ and various numbers of elements (*N*) are presented in Figure 11. The dashed red line indicates the reference level corresponding to NNetEn with $N = 19625$. A decrease in *N* below 19,625 should not lead to a significant change in entropy relative to the reference level. The method of entropy measurement is most stable for $N \geq 11{,}000$, when the NNetEn values almost coincide with the reference level. For $N < 11{,}000$, the dependence of NNetEn on *N* is observed, and it is reflected in the appearance of fluctuations in NNetEn relative to the reference level.

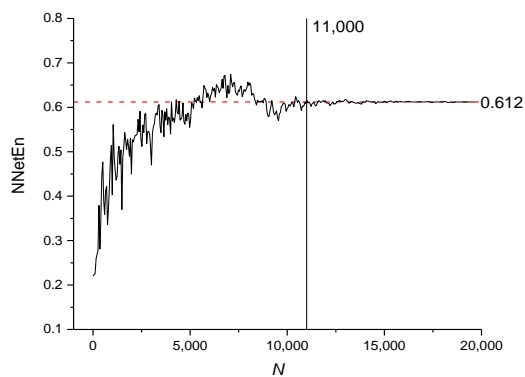

**Figure 11.** The effect of time series length on NNetEn values for time series based on a logistic map (Equation (2)).

The NNetEn values for the sine periodic map (Equation (7)) and binary map (Equation (8)) with various number of elements *N* are presented in Figures 12 and 13, respectively.

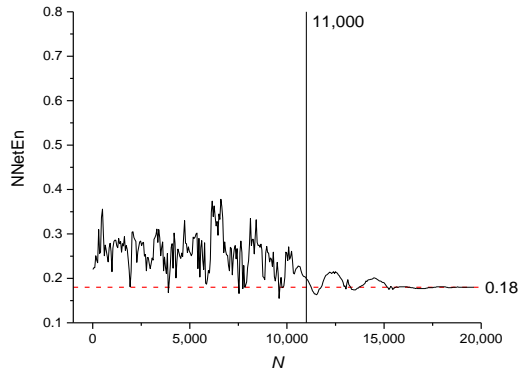

**Figure 12.** The effect of time series length on NNetEn values for sine periodic time series (Equation (7)).

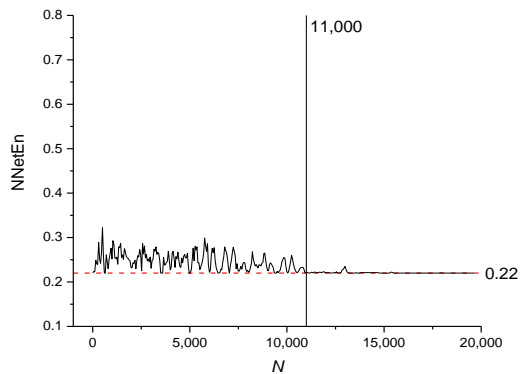

**Figure 13.** The effect of time series length on NNetEn for binary time series (Equation (8)).

It can be concluded that for all types of signals considered, the most stable result is obtained for $N > 11{,}000$.

*3.5. A Comparison between the NNetEn Measure and Other Entropy Methods*

A comparison between NNetEn (number of epochs is 100) and the ApEn entropy values for the logistic map time series at $N = 19{,}625$ is depicted in Figure 14. Variables *rr* and *m* are the main parameters of ApEn [10]. The dependencies of NNetEn and ApEn on *r* almost repeat each other, except for some features. To describe these features, the following labels have been introduced: $r_1 = 3.60666667$, $r_2 = 3.68833333$, $r_3 = 3.835$, and $r_4 = 3.94833333$. The value of entropy has a local minimum for NNetEn and ApEn (*rr* = 0.025, *m* = 2) at $r = r_1$, while ApEn (*rr* = 0.1, *m* = 2) reaches a local maximum at this point. This demonstrates that ApEn is very sensitive in its parameters (*rr* and *m*). The proposed method resolves this problem, and the position of the minimum does not depend on the parameters of the method. In addition, the proposed method is compared with other entropy measures (Table 1). All entropies have an increasing *r* trend, and the entropy at $r = r_1$ is less than the entropy at $r = r_4$. All entropies have a minimum of $r = r_1$ and a maximum of $r = r_4$. Therefore, the NNetEn entropy measure gives similar results to other methods for calculating entropy, while having the advantage of a small number of parameters and a high computational stability, independent of the signal amplitude and the length of the time series.

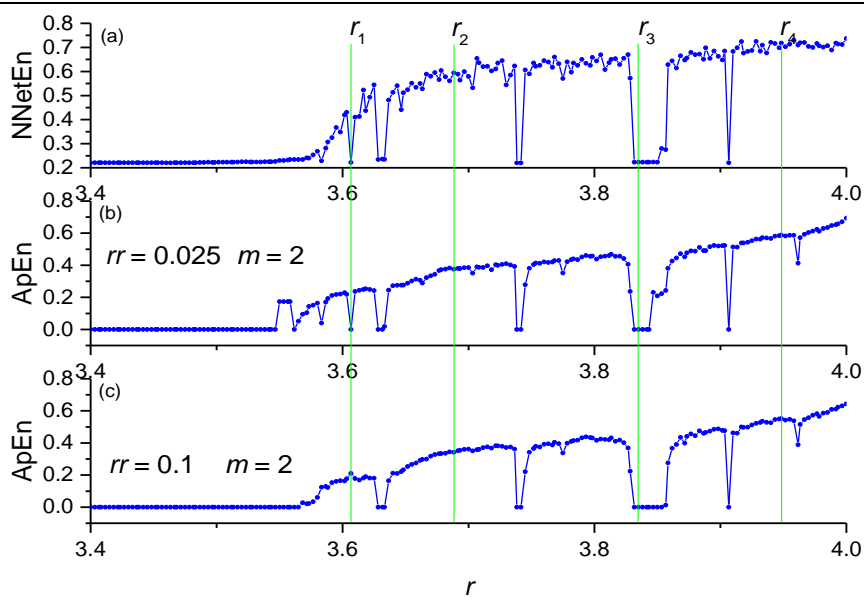

**Figure 14.** A comparison between NNetEn and ApEn for the logistic mapping. NNetEn (number of epochs is 100) (**a**); ApEn ($rr$ = 0.025, $m$ = 2) (**b**); ApEn ($rr$ = 0.1, $m$ = 2) (**c**).

**Table 1.** Comparison between NnetEn and other entropy measures.

|  | $r_1$ = 3.60666667 | $r_2$ = 3.68833333 | $r_3$ = 3.835 | $r_4$ = 3.94833333 |
|---|---|---|---|---|
| NNetEn Entropy | 0.2208 | 0.6193 | 0.2222 | 0.6928 |
| ApEn [10], $rr$ = 0.025, $m$ = 2 | 0 | 0.374 | 0 | 0.588 |
| ApEn [10], $rr$ = 0.1, $m$ = 2 | 0.210 | 0.341 | 0 | 0.552 |
| Topological entropy [24] | 1.16 | 1.21 | 0.61 | 1.21 |
| SampEn [1] | 0.067 | 0.287 | 0 | 0.49 |
| PerEn [1] | 0.45 | 0.53 | 0.347 | 0.71 |
| DispEn [1] | 0.45 | 0.649 | 0.308 | 0.683 |
| Shannon entropy from recurrence plots [8] | 0.85 | 1.79 | 0.1 | 2.29 |
| Shannon entropy from symbolic dynamics [8] | 2.45 | 4.81 | 1.15 | 6.97 |

## 4. Discussion

Investigating the complexity of non-linear time series has become a hot topic in recent years, as it has many practical applications. Various entropy measures have been introduced to calculate the complexity of a given time series. However, almost all existing methods are based on Shannon entropy or conditional probability, which suffers from a sensitivity to the length of the time series and its initial parameters. We propose the NNetEn entropy measure to overcome these difficulties. NNetEn is an estimation method based on MNIST-10 classification using the LogNNet 784:25:10 neural network. Its calculation algorithm is different from that of other existing methods. The NNetEn algorithm needs a reservoir matrix with 19,625 elements filled with time series. As many practical applications use a time series with $N < 19,625$ elements, we propose a method for filling the matrix with $N < 19,625$. The chosen method provides the most stable results for calculating NNetEn in comparison to similar methods of filling the matrix by rows or by stretching a row.

In future studies, it would be possible to investigate other network configurations with a different number of neurons, and it could lead to a change in the number of elements in the reservoir matrix. Nevertheless, the presented method for the use of LogNNet 784:25:10 has promising opportunities for practical application and may be of interest to the scientific community.

Figures 11–13 demonstrate that there are entropy modulations due to changes in time series length in the matrix filling method. This is caused by shifts in the time series relative to the rows in the $W_1$ matrix.

This method is most stable at $N \geq 11,000$, when the NNetEn values almost coincide with the reference level. In practice, when comparing time series, we recommend using time series of the same length and, if possible, $N \geq 11,000$.

The proposed method is applied on constant, binary, periodic, and various chaotic time series. The results demonstrate that the NNetEn value lies between 0.1 and 1. The lower limit of NNetEn is 0.1, as the minimum classification accuracy is 10%. This is achieved when images of digits are recognized from 10 random options.

The NNetEn value converges on a plateau with an increase in the number of epochs. The number of epochs is considered the input parameter of the method. The behavior of NNetEn values is roughly similar for the logistic map when the number of epochs is 100 and when the number of epochs is 400. Therefore, we use 100 epochs in LogNNet and suggest using at least 100 epochs in other examples.

In Section 3.3, the parameter $\delta_{Ep1/Ep2}$ is introduced to compare the effect of the number of epochs on NNetEn values. This parameter can be considered a new characteristic of the input time series. We call it the learning inertia of the time series. This parameter characterizes the speed of training of a network for the given time series. Small values of $\delta_{Ep1/Ep2}$ correspond to the rapid achievement of a plateau by NNetEn values with an increasing the number of epochs. The parameter $\delta_{Ep1/Ep2}$ may depend on the initial distribution of the matrix $W_2$ elements or on the learning rate of the back propagation method. Figure 10 shows the bursts at the border of the order-chaos regions. Further study of the learning inertia of time series for various signals and the study of transitions from order to chaos may become topics for further research.

The proposed model has the following advantages compared to the existing entropy measurement methods:

- It is simple to use.
- It has one control parameter—the number of epochs—when training the network.
- Scaling the time series by amplitude does not affect the value of entropy.
- It can be used for a series of any length. The most repeatable results are observed when $N$ varies in the range $N = 11,000$–19,626.
- It outperforms the existing methods.
- A new characteristic of the time series is introduced—learning inertia. This can be used to identify additional patterns in the dynamics of the time series.

For example, the study [1] introduced a method of entropy estimation depending on the embedded dimension *m*, time delay *d*, and constant value *c*. Any changes in these parameters lead to different results. In contrast, the proposed method depends only on the number of epochs, and the position of the minima and maxima does not depend on the number of epochs.

Extending the current study to multivariable time series can be considered as another direction for future work. In addition, it would be beneficial to apply the method of calculating NNetEn in practice to process data from medical, physical, biological, and geophysical experiments. The use of neural networks for calculating entropy and other characteristics of time series could become a promising direction for further research.

**5. Conclusions**

This study proposes a new entropy measure called NNetEn for evaluating the complexity of the given time series. NNetEn is the first entropy measure that is based on artificial intelligence methods. The method modifies the structure of the LogNNet classification model so that the classification accuracy of the MNIST-10 digits dataset indicates the degree of complexity of a given time series. The calculation results of the proposed model are similar to those of existing methods, while the model structure is completely different and provides considerable advantages. For example, it overcomes such difficulties as the method's sensitivity to the length and amplitude of the time series. The method has only one input parameter and is easier to use, which is important in practical applications. In addition to the method for measuring entropy, an equation for calculating a new characteristic of a time series, learning inertia, is given. The study results can be widely applied in practice and should be of interest to the scientific community.

**Supplementary Materials:** The following are available online at https://www.mdpi.com/article/10.3390/e23111432/s1

**Author Contributions:** Conceptualization, A.V.; methodology, A.V. and H.H.; software, A.V.; validation, H.H. and A.V.; formal analysis, H.H.; investigation, A.V.; resources, H.H.; data curation, H.H.; writing—original draft preparation, H.H. and A.V.; writing—review and editing, A.V. and H.H.; visualization, A.V.; supervision, A.V.; project administration, A.V.; funding acquisition, A.V. All authors have read and agreed to the published version of the manuscript.

**Funding:** This research received no external funding.

**Institutional Review Board Statement:** Ethical review and approval were waived for this study.

**Informed Consent Statement:** Not applicable.

**Data Availability Statement:** The database of handwritten digits MNIST-10 (available on Yan LeCun's Internet page [25]) was used for the study.

**Acknowledgments:** The authors express their gratitude to Andrei Rikkiev for valuable comments made in the course of the article's translation and revision.

**Conflicts of Interest:** The authors declare no conflicts of interest.